\documentclass[runningheads]{llncs}

\usepackage[utf8]{inputenc}
\usepackage[T1]{fontenc}
\usepackage{graphicx}
\usepackage{amsmath,amssymb}
\usepackage{booktabs}
\usepackage{algorithm}
\usepackage{algorithmic}
\usepackage{cite}
\usepackage{hyperref}
\usepackage{fancyhdr}

\title{Not every day is a sunny day: Synthetic cloud injection for deep land cover segmentation robustness evaluation across data sources}
\titlerunning{Synthetic cloud injection for land cover segmentation}

\author{Sara Mobsite\inst{1} \and
        Renaud Hostache\inst{1} \and
        Laure Berti-\'Equille\inst{1} \and
        Emmanuel Roux\inst{1} \and
        Joris Gu\'erin\inst{1}} 
\institute{1 ESPACE-DEV, French National Research Institute for Sustainable Development (IRD)\\
\email{\{firstname.lastname\}@ird.fr}}

\begin{document}

\maketitle

\thispagestyle{fancy}
\pagestyle{fancy}

\begin{abstract}
Supervised deep learning for land cover semantic segmentation (LCS) relies on labeled satellite data. However, most existing Sentinel-2 datasets are cloud-free, which limits their usefulness in tropical regions where clouds are common. To properly evaluate the extent of this problem, we developed a cloud injection algorithm that simulates realistic cloud cover, allowing us to test how Sentinel-1 radar data can fill in the gaps caused by cloud-obstructed optical imagery. We also tackle the issue of losing spatial and/or spectral details during encoder downsampling in deep networks. To mitigate this loss, we propose a lightweight method that injects Normalized Difference Indices (NDIs) into the final decoding layers, enabling the model to retain key spatial features with minimal additional computation. Injecting NDIs enhanced land cover segmentation performance on the DFC2020 dataset, yielding improvements of 1.99\% for U-Net and 2.78\% for DeepLabV3 on cloud-free imagery. Under cloud-covered conditions, incorporating Sentinel-1 data led to significant performance gains across all models compared to using optical data alone, highlighting the effectiveness of radar-optical fusion in challenging atmospheric scenarios. The code for our cloud location mask generation algorithm is publicly available at: \href{https://github.com/sara-mobsite/cloud-generation}{code}.

\end{abstract}

\section{Introduction} 

High-resolution land cover segmentation (LCS) increasingly relies on deep neural networks, which facilitate precise pixel-wise classification of satellite imagery by learning from labeled datasets.  LCS can be utilized in environmental studies to track deforestation \cite{choi2024deforestation}, identify forest disturbances \cite{tian2025vision}, and map the coverage of sea ice \cite{wang2025lh}, among other applications. The type of satellite data selected for a given application is guided by the specific objectives and requirements of each use case. Some studies leverage Synthetic Aperture Radar (SAR) data, such as Sentinel-1  \cite{jamali2024residual, catry2018wetlands, pech2024segmentation}, while others utilize optical imagery from satellites such as Sentinel-2 \cite{buttar2024generating, andresini2024leveraging}. SAR and optical data are increasingly combined in research due to their complementary properties \cite{fuller2023croma, tzepkenlis2023efficient}: optical imagery provides rich textural and spatial details, while SAR data is almost unaffected by clouds and adverse weather conditions.

In deep learning-based LCS, SAR data provides complementary information that can help compensate for gaps in optical imagery caused by atmospheric interference, such as clouds. However, existing datasets that combine Sentinel-1 and Sentinel-2 data, such as DFC2020 \cite{IEEE2020DataFusion}, typically exclude cloud-covered samples and do not include a dedicated cloud class. This limitation is significant, as cloud presence is common in real-world satellite observations, especially in tropical and subtropical regions \cite{zhang2021global}. The lack of optical images containing cloud cover in training samples can hinder generalization and reduce segmentation accuracy in real-world applications. Moreover, this limits the model’s ability to learn how to effectively utilize SAR features to compensate for missing optical information during the training process. 

In this work, we propose to compare the performance of state-of-the-art segmentation networks while assessing the influence of the input satellite data type (SAR, optical, or a combination of both) and the effect of clouds. To remain in a controlled environment, we propose a more realistic evaluation approach for comparing LCS networks by injecting synthetic clouds into originally cloud-free optical data during both the training and testing processes. This approach helped leverage labeled optical data and make it more suitable for real-world applications. To do so, we introduce an automated method that generates cloud masks based on real cloud patterns and insert them into Sentinel-2 imagery.   Our goal is therefore to evaluate the impact of cloud cover on segmentation performance and to assess the ability of SAR features to compensate for cloud-induced gaps.

One of the challenges with autoencoder-based networks for LCS is the downsampling applied during the encoding stage. This downsampling often results in the loss of spatial details and features necessary for accurately identifying and distinguishing land cover classes at high resolution. Therefore, in this work, we explore a second approach aimed at enhancing the spatial and textural details in optical data under minimal cloud coverage. To mitigate the loss of fine features caused by downsampling in the encoder stage, we employ a decoder-level feature injection strategy using normalized difference indices (NDIs). Specifically, NDIs for water (NDWI), buildings (NDBI), and vegetation (NDVI) are computed and concatenated with decoder features at the final segmentation layer. This approach reintroduces important spatial and textural cues associated with key land cover types, thereby improving the model’s ability to distinguish between land cover classes. 

\section{Related Work}

LCS heavily relies on spectral features that provide reflectance values across different wavelengths, as well as spatial features that capture patterns and textures based on pixel arrangements. Sentinel-2 offers rich detail but is hindered by cloud and atmospheric interference, while Sentinel-1 is weather-resilient. Consequently, many studies explore the fusion of data from both sources using early, middle, or late-level fusion. In early-level fusion, data is combined at the input stage before encoding.  In \cite{guisao2023forest}, the authors focused on forest identification in the Bajo Cauca Subregion of Colombia by applying early-stage fusion. They proposed a Random Forest algorithm that uses dual-polarization C-band SAR data, combined with vegetation indices extracted from multispectral optical data. In \cite{tzepkenlis2023efficient}, a modified U-TAE model based on channel attention was proposed, using different combinations of satellite data at the input level, including spectral indices and elevation data, to improve LCS performance.

In \cite{fuller2023croma}, feature-level fusion was applied by independently encoding SAR and optical data using Vision Transformer encoders, followed by a second Vision Transformer encoder that performed the fusion of the extracted features. In \cite{ma2022amm}, attention-based fusion at the channel level was used to learn from both SAR and optical data. The attention-based multi-modal network combined features from both modalities by learning channel-wise relationships between the outputs of the multi-scale feature extractors.  In \cite{khankeshizadeh2024fba}, a multi-stream U-Net architecture was proposed to detect forest burned areas, incorporating residual and stream-to-stream connections for feature fusion. The fusion process was based on multi-level integration of features extracted from two independent encoding branches. These feature maps were then passed into a unified decoder, where the combined outputs from each encoder stream were reintroduced through residual connections to enhance spatial feature recovery after downsampling.   Rather than relying solely on feature-level fusion, the authors in \cite{gargiulo2020integration} implemented fusion at the final stage. In their approach, each processing stream handled optical and SAR data separately, and the final pixel-wise classification was derived by combining the feature maps generated by each autoencoder. Consequently, instead of using a U-Net architecture, the study proposed a W-Net network. 

Regardless of the fusion technique used, most available multi-modal datasets for supervised LCS only include cloud-free images. This does not reflect real-world conditions, especially for Sentinel-2 imagery, which is highly susceptible to cloud cover and other atmospheric disturbances. This omission presents a key gap in the literature, restricting insight into how effectively SAR features can compensate for cloud-related gaps in optical data.

\section{Method} 

\subsection{SAR and Optical Data Fusion Architecture}

Our work aims to identify the most effective and reliable data sources for high-resolution LCS. We evaluate three input configurations: optical data alone, SAR data alone, and a combination of both at the input level. This setup enables a comprehensive assessment of segmentation performance for each data modality. For this analysis, we use two state-of-the-art neural network architectures for semantic segmentation: U-Net \cite{ronneberger2015u},  and DeepLabV3 \cite{chen2017rethinking}.

\subsection{ Improving Segmentation with NDIs}
\begin{figure}
  \centering
  \includegraphics[width=0.7\linewidth]{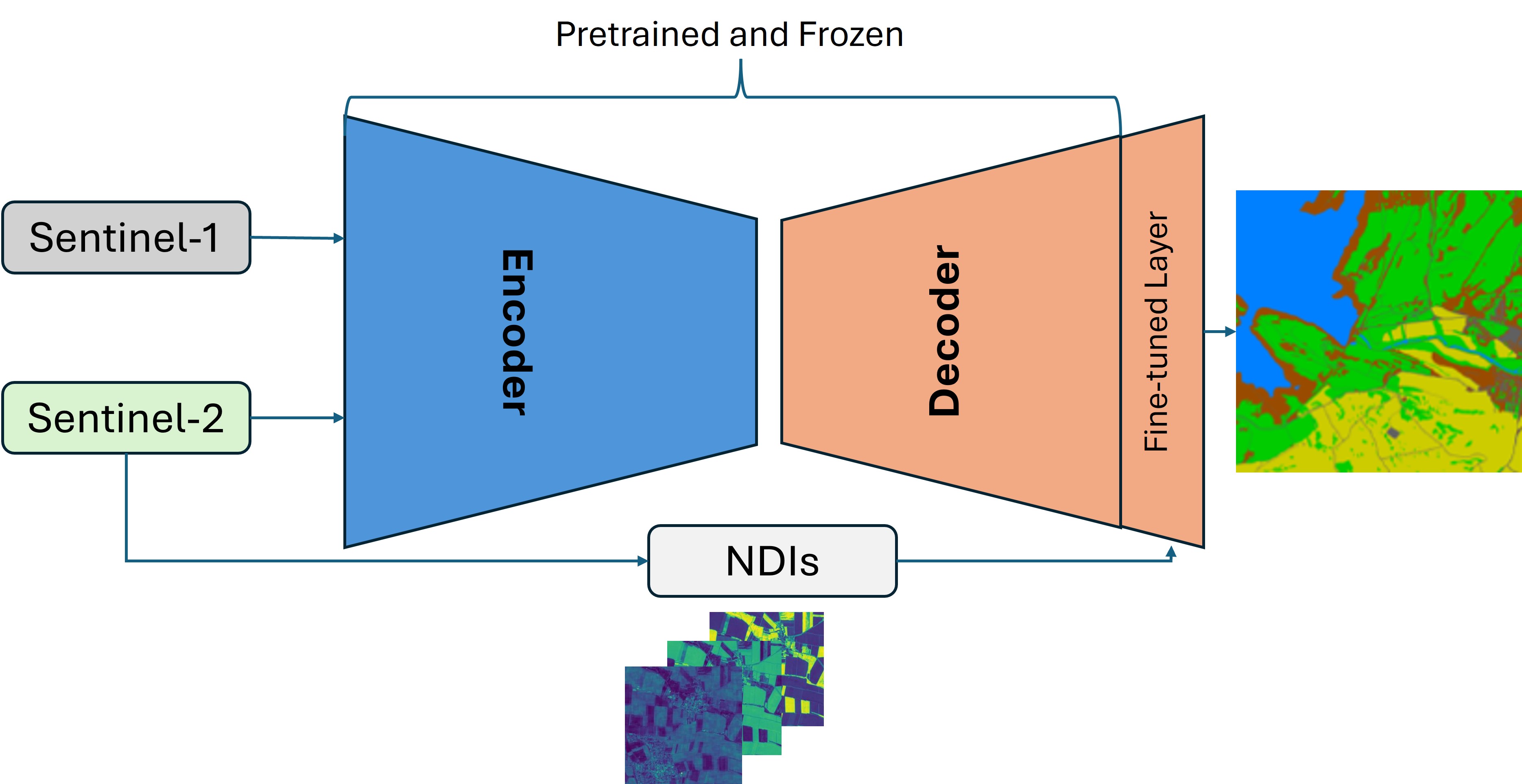}
\caption{NDIs injection  in the decoder of a pretrained autoencoder } 
  \label{fig:short}
\end{figure}

Sentinel-1  provides dual-polarization based data, including VV (vertical transmit and vertical receive) and VH (vertical transmit and horizontal receive) modes. Sentinel-2 data, on the other hand, provides extensive spectral coverage through its 13 bands spanning the visible, near-infrared (NIR), and shortwave-infrared (SWIR) ranges.

In segmentation networks, the encoder stage often leads to the loss of important spatial details due to repeated downsampling. To mitigate this issue, our approach incorporates normalized difference indices (NDIs), including vegetation, water, and built-up area indices. These NDIs are injected into the later stages of the decoder to enhance spatial feature recovery across the studied networks. We specifically selected these indices to be reintroduced in the final layers to reinforce information related to the primary land-cover target classes (water, built-up areas, and vegetation).

As illustrated in Fig. 1, the networks are initially trained on the DFC2020 dataset using either optical data alone or a fusion of optical and SAR data, without including the NDIs. In the second stage, transfer learning is applied: all previously trained layers are frozen, and only the final 1$\times$1 convolutional layer is retrained. At this stage, the output feature maps from the decoder (before the final layer) are concatenated with the NDIs. This combined feature representation is then passed through a 3×3 convolutional layer followed by a final 1×1 convolution to perform LCS.

 In our study, we inject three widely used NDIs in the decoding final stage: the Normalized Difference Vegetation Index (NDVI) \cite{pettorelli2013normalized}, the Normalized Difference Water Index (NDWI) \cite{gao1996ndwi}, and the Normalized Difference Built-up Index (NDBI) \cite{zha2003use}. These indices were selected because they are specifically designed to highlight key land cover types, such as vegetation, water bodies, and built-up areas, which are essential for accurate LCS. In \cite{wang2024feature}, these indices were used to train a self-supervised network based on masked autoencoders, encouraging the model to learn important features from optical data.  Instead of using raw optical data, these indices provide a concise representation of key land cover types using only three channels instead of all 13 spectral inputs. This makes the approach more efficient and helps emphasize relevant spatial features for the main land cover classes. 
\subsection{Synthetic Cloud Generation and Injection into Optical Data}

Existing datasets based on Sentinel-2 are mostly cloud-free,  which does not accurately represent real-world conditions. To address this, we developed an algorithm that synthetically injects cloud patterns into clean Sentinel-2 images.  Rather than manually generating cloud-covered samples by downloading cloud data for each image from open-access sources, we create unique, cloud-like patterns by generating binary cloud location masks and embedding real-world cloud radiometric data into them. These binary cloud location masks are generated by randomly shaping ellipses based on a normal Gaussian distribution and bilinear interpolation, which mimics the irregular appearance of natural clouds (Algorithm 1). In our approach, the deformation, density, and cloud coverage percentages are uniquely defined for each mask. This helps prevent repetitive patterns in Sentinel-2 data, which can lead to overfitting during the learning process.

\begin{figure}[H]

  \centering
  \includegraphics[width=0.99\linewidth]{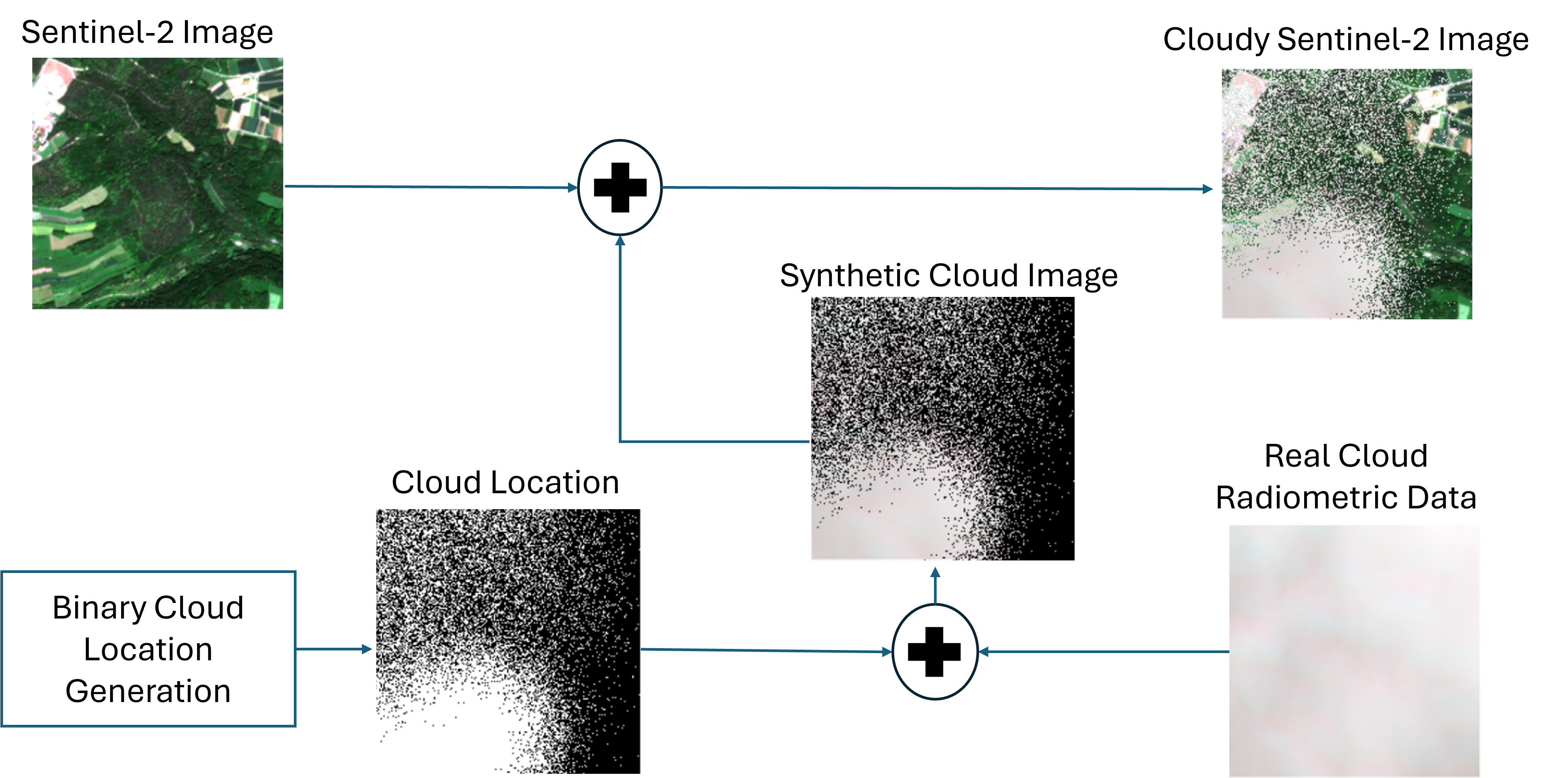}
\caption{ Example of synthetic cloud  generation based on a binary cloud location  mask  and cloud  optical data} 
  \label{fig:short}
\end{figure}
 The cloudy areas of the mask are filled with real thick cloud radiometric values, while clear sky regions preserve the original Sentinel-2 imagery targeted for modification.  Sentinel-1 was not modified, as it is not affected by cloud cover. This method allows for scalable and rather realistic injection of clouds into cloud-free datasets, supporting the training of more robust models capable of handling diverse atmospheric conditions.

\begin{algorithm}[H]
\caption{Cloud Location Mask Generation}
\begin{algorithmic}[1]
\REQUIRE Number of masks $N$, mask size $S \times S$, max coverage fraction $C_{\text{max}}$, ellipse axes bounds $E_{\min}, E_{\max}$, deformation range $D_{\text{max}}$
\FOR{$i = 1$ to $N$}
    \STATE Initialize empty binary mask $M \in \mathbb{R}^{S \times S}$
    \STATE Calculate the number  of cloud pixels $T \leftarrow S^2 \cdot \text{random}~C \in [0, C_{\text{max}}]$
    \STATE Set filled area $A \leftarrow 0$
    \WHILE{$A < T$}
        \STATE Create blank image $B \in \mathbb{R}^{S \times S}$
        \STATE Draw filled ellipse on $B$ with:
        \begin{itemize}
            \item Random center within bounds
            \item \text{random}~Axes $\in [E_{\min}, E_{\max}]$
            \item \text{random}~Angle $\in [0^\circ, 180^\circ]$
        \end{itemize}
        \STATE Generate random flow field $F \in \mathbb{R}^{S \times S \times 2}$:
        \begin{itemize}
            \item Each component $\sim \mathcal{N}(0,1)$
            \item Multiply by random $D \in [5, D_{\text{max}}]$
        \end{itemize}
        \STATE Deform $B$ using $F$ via bilinear remapping (reflect border) $\rightarrow B_{\text{def}}$
        \STATE Threshold $B_{\text{def}}$ to binary $\rightarrow B_{\text{bin}}$
        \STATE Merge: $M \leftarrow M \lor B_{\text{bin}}$
        \STATE Update $A \leftarrow \text{countNonZero}(M)$
    \ENDWHILE
    \STATE Save $M$
\ENDFOR
\end{algorithmic}
\end{algorithm}

Fig. 2 presents an example of a binary cloud location mask generated randomly based on Algorithm 1, alongside the resulting cloud mask after injecting radiometric values of real cloud data derived from Sentinel-2  \textbf{100\% cloud coverage} observation. The cloud data is extracted from a tile captured on \textbf{May 19, 2025, at 14:27:19 UTC}, covering a $100\,\text{km} \times 100\,\text{km}$ region in northeastern Brazil, centered near \textbf{3\textdegree{} south latitude} and \textbf{44\textdegree{} west longitude}, within the state of Maranhão.  In particular, this Sentinel-2 image is used in our work solely as a source of real cloud data. The region is divided into \textbf{1,500 non-overlapping tiles} of \textbf{256 $\times$ 256 pixels}, and each tile is used to generate a unique cloud mask. To simulate realistic atmospheric interference, the generated cloud masks are individually applied to each optical image in the DFC2020 dataset, ensuring unique cloud coverage and patterns per sample.

\section{Experiments}
\subsection{Dataset}
The DFC2020 dataset, introduced for the IEEE GRSS Data Fusion Contest 2020 as an extension of the SEN12MS dataset \cite{schmitt2019sen12ms, IEEE2020DataFusion}, contains  Sentinel-1 and Sentinel-2 image pairs along with pixel-level land cover annotations at a resolution of 10 meters. The dataset covers 7 land cover classes. Following prior studies \cite{fuller2023croma, wang2024feature}, we used the original test split (5,128 samples) for model training (with an 85\%/15\% train/validation split), and the original validation set (986 samples) for final evaluation. In addition to the clean dataset, we evaluate the robustness of different approaches using a modified version with simulated cloud cover. To reflect realistic atmospheric conditions, clouds were synthetically injected into both the training and evaluation samples. Each image was overlaid with a unique, generalized cloud mask to ensure diverse and realistic cloud patterns. The deformation applied to the binary cloud location masks varied between 5 and 20 pixels, the random axes ranged from 10 to 100, and the simulated cloud coverage ranged from 0\% to 70\% per sample.

\subsection{Implementation Details}

Each of the proposed networks was based on a ResNet-50  \cite{he2016deep} backbone as an encoder, with weights initialized from a pretrained model on the ReBEN dataset \cite{clasen2024reben} for land cover classification. The models were implemented using the PyTorch library and trained on an NVIDIA RTX 4000 GPU. The input size for the Sentinel images was set to 256×256 pixels. For all experiments, cross-entropy loss was used as the objective function. Models were trained for up to 200 epochs, with early stopping applied using a patience of 20 epochs. The Adam optimizer was used with an initial learning rate of 0.001, and the batch size was set to 32. For the NDIs injection phase, the models were trained using the same configuration.

\section{Results and Discussion}

\subsection{The Impact of NDIs Decoder-Level Injection on Model Performance}
Table 1 shows the results of each network with and without NDIs injection. The networks were first trained without this operation. Then, pretrained weights were transferred, and NDIs were injected into the final decoder layer, while all earlier layers were frozen to train only the NDI-related part. This approach avoids full retraining and focuses on enhancing the decoder. Adding NDIs improved the U-Net’s mean IoU by 0.59\% with optical input alone and by 2.00 with SAR+Optical input. For DeepLabV3, the gains were 5.18 and 2.78, respectively, showing a stronger benefit, especially with optical-only input. This larger impact on DeepLabV3 is likely due to its architecture, which relies on an encoder-decoder structure with upsampling that lacks skip connections from early layers of the encoder. In contrast, U-Net incorporates skip connections from early encoding layers, which already preserve spatial details, making the addition of NDIs a more complementary enhancement rather than a critical compensatory input.
\begin{table}[htbp]
  \centering
  \caption{Mean IoU of Each Model Using Different Input Types on the Cloud-Free DFC2020 Test Set, Including  Performance of Decoder-Level NDIs Injection.}
  \begin{tabular}{l|c|c|c|c|c}
    \toprule
    \textbf{Model} 
      & \textbf{SAR} 
      & \textbf{Optical} 
    & \textbf{Optical+NDIs} 
      & \textbf{SAR+Optical} 
      & \textbf{SAR+Optical+NDIs} \\
    \midrule
    U-Net       
      & 38.89           
      & 47.80     
      & 48.39 (+0.59\%)  
      & 51.62           
      & 53.61 (+1.99\%)  \\

    DeepLabV3   
      & 36.88           
      & 39.24             
      & 44.42 (+5.18\%) 
      & 43.22 
      & 46.00 (+2.78\%)  \\
    \bottomrule
  \end{tabular}
  \small
\end{table}

\subsection{The Impact of  Cloud Injection on Model Performance}

To assess the impact of using SAR data alone or in combination with optical data under cloudy conditions, we evaluated the performance of each model on cloud-injected data. The training was conducted under two scenarios: using cloud-free data and using data with injected clouds. The results show that training with cloud-injected data (TCI) significantly improves performance when using optical inputs alone. For instance, DeepLabV3's mean IoU rises from 15.82 to 43.48, and U-Net’s from 15.44 to 28.36, highlighting that without exposure to cloudy conditions during training, models struggle to handle occlusions in the test set. When using SAR data alone, the performance remains stable because SAR is cloud-penetrating and unaffected by optical obstructions. However, combining SAR with optical data only improves performance when the model is trained with cloudy data. For example, DeepLabV3  without TCI sees a drop to 14.16 with SAR+Optical (worse than using SAR alone), while training with TCI boosts it to 45.71, showing that SAR is only useful for filling gaps if the model learns to fuse SAR and optical information under cloud-obscured conditions. This suggests that SAR data does not inherently compensate for clouds unless the model is explicitly trained to associate SAR with missing optical context.

\begin{table}[htbp]
  \centering
  \caption{Mean IoU of Each Optical-Based Model Trained With  Cloud Injection (TCI) and Without it. Evaluated on the Cloud-Contaminated DFC2020 Test Set}
  \begin{tabular}{l|c|c|c|c|c}
    \toprule
    \textbf{Model} 
 
      & \textbf{Optical} 
      & \textbf{Optical TCI} 
      & \textbf{SAR}
      & \textbf{SAR+Optical} 
      & \textbf{SAR+Optical TCI} \\
    \midrule
    U-Net       &     15.44      &   28.36        &  38.89            &      17.44   &     47.72   \\
    DeepLabV3    &         15.82    &    43.48    &          36.88  &          14.16  &         45.71    \\
    \bottomrule
  \end{tabular}
  \small
\end{table}

\begin{table}[htbp]
  \centering
  \caption{Mean IoU of Each Optical-Based Model Trained With  Cloud Injection (TCI) and Without it. Evaluated on the Cloud-free DFC2020 Test Set}
  \begin{tabular}{l|c|c|c|c|c}
    \toprule
    \textbf{Model} 
 
      & \textbf{Optical} 
      & \textbf{Optical TCI} 
      & \textbf{SAR}
      & \textbf{SAR+Optical} 
      & \textbf{SAR+Optical TCI} \\
    \midrule
    U-Net       &     47.8    &     35.42     &  38.89            &   51.62    &    51.26  \\
    DeepLabV3    &        39.24    &   45.64  &          36.88  &    43.22        &   45.66    \\
    \bottomrule
  \end{tabular}
  \small
\end{table}

\section*{Conclusion}
This study demonstrates the importance of training with cloud-injected data to enable models to effectively handle cloud occlusion in remote sensing imagery. SAR data alone offers robust performance under cloudy conditions, but its full potential is only realized when combined with optical data in models specifically trained to fuse both modalities under such conditions.  Even optical-only models benefit from cloud-injected training by learning to infer missing data from nearby clear regions. Decoder-level NDI injection further enhances spatial detail recovery, particularly in DeepLabV3, by facilitating the reconstruction of features lost during downsampling. Future work could explore cloud synthesis using generative deep learning networks and assess performance based on cloud coverage percentage in SAR-optical fusion tasks.

\section*{Acknowledgement}
This work has received funding from the European Union’s Horizon Europe Research and Innovation program under Grant Agreement Nº 101137398

\bibliographystyle{plain}

\bibliography{bibliography}
\end{document}